\documentclass[10pt,twocolumn,letterpaper]{article}

\usepackage{cvpr}
\usepackage{times}
\usepackage{epsfig}
\usepackage{graphicx}
\usepackage{amsmath}
\usepackage{amssymb}
\usepackage{multirow}
\usepackage{booktabs}

\usepackage[pagebackref=true,breaklinks=true,letterpaper=true,colorlinks,bookmarks=false]{hyperref}

\cvprfinalcopy 

\def\cvprPaperID{****} 

\ifcvprfinal\fi
\begin{document}

\title{TransAction: ICL-SJTU Submission to EPIC-Kitchens Action Anticipation Challenge 2021}

\author{Xiao Gu$^{1}$, Jianing Qiu$^{1}$, Yao Guo$^{2}$, Benny Lo$^{1}$, Guang-Zhong Yang$^{2}$\\
$^{1}$Imperial College London, UK\\
$^{2}$Shanghai Jiao Tong University, China\\
{\tt\small \{xiao.gu17,jianing.qiu17,benny.lo\}@imperial.ac.uk, \{yao.guo, gzyang\}@sjtu.edu.cn}
}

\maketitle

\begin{abstract}
In this report, the technical details of our submission to the EPIC-Kitchens Action Anticipation Challenge 2021 are given. We developed a hierarchical attention model for action anticipation, which leverages Transformer-based attention mechanism to aggregate features across temporal dimension, modalities, symbiotic branches respectively. In terms of Mean Top-5 Recall of action, our submission with team name ICL-SJTU achieved $13.39\%$ for overall testing set, $10.05\%$ for unseen subsets and $11.88\%$ for tailed subsets. Additionally, it is noteworthy that our submission ranked 1st in terms of verb class in all three (sub)sets.       
\end{abstract}

\section{Introduction}
Egocentric action anticipation \cite{damen2020epic} is receiving increasing attention recently, which aims to anticipate what the subject to do next based on the recordings from egocentric cameras. Different from the third-person action anticipation, it actually records what the subject observes and performs high-level perception of in the brain. Associating past sensory input with future actions is a fundamental step for understanding human cognition mechanisms. 

It is a challenging problem since future events are highly uncertain, and there exist several possible diverse predictions based on the observation of the past \cite{furnari2018leveraging}. It is difficult to establish an explicit model between the past and the future, as the sensory input (e.g. visual observation) may have asynchronous casual effect on the next action and the future is of multi-modality in nature. Directly arranging the sensory input as a sequential order and feeding it to some conventional temporal modelling architectures (e.g. RNN) may tend to ignore the effects contributed by some relatively old experiences. In our submission, we adopted the Transformer to dynamically fuse information across time, modalities, and \textit{verb} \& \textit{noun} branches. 

\begin{figure}[tp]
    \centering
    \includegraphics[width=\linewidth]{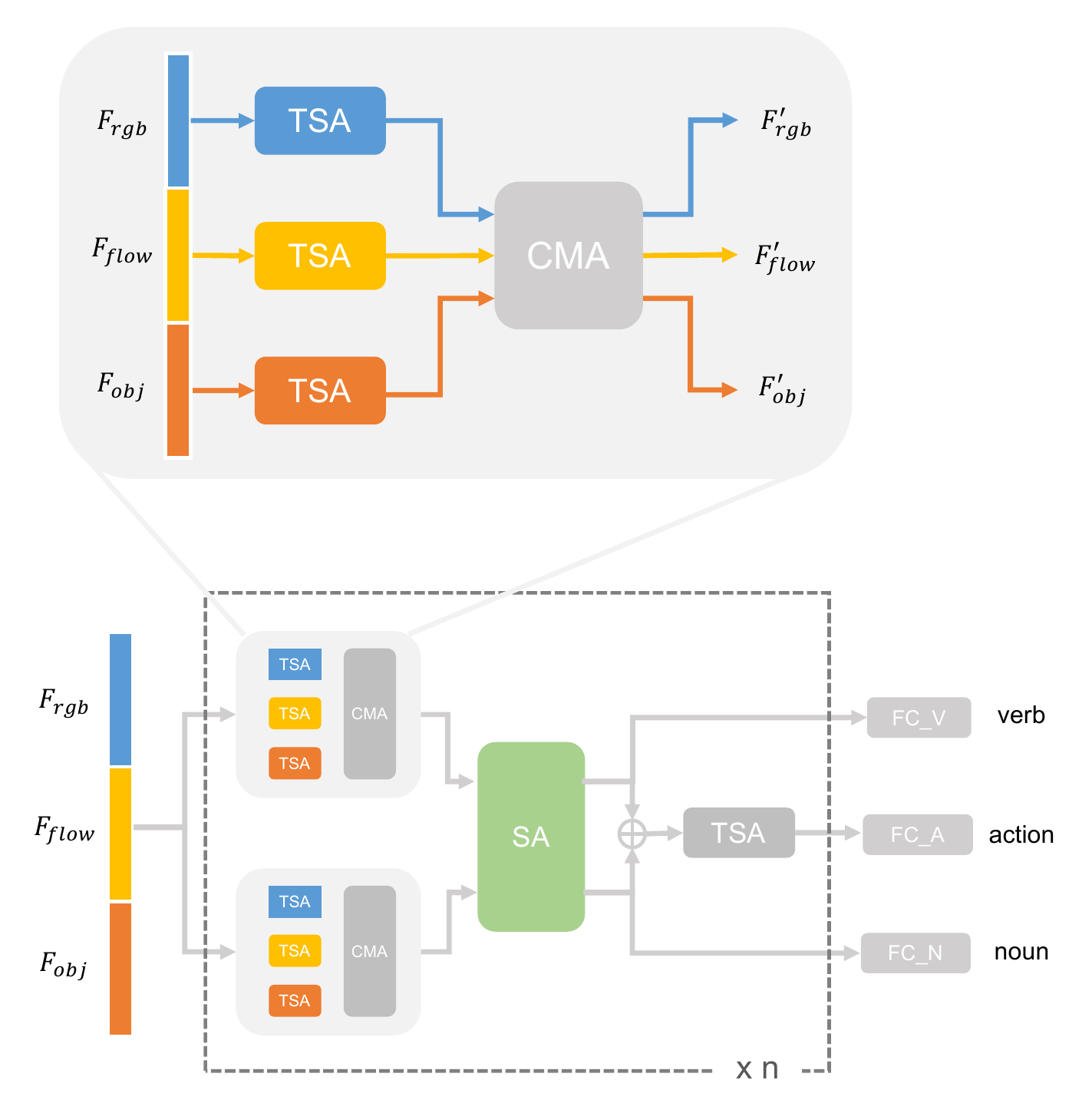}
    \caption{Overview of our hierarchical Transformer-based fusion framework. Our framework is a cascade of several singular blocks. In each block, the temporal self-attention (TSA) module aims to model long-range temporal information, capturing asynchronous effect for the action anticipation. The cross-modality attention (CMA) module aims to fuse information across modalities via Transformer-based attention mechanism. The symbiotic attention (SA) module serves for the mutual interaction between \textit{verb} and \textit{noun} branches with the goal of benefiting each other.}
    \label{fig:framework}
\end{figure}

On the other hand, each label of egocentric actions in Epic-Kitchen is formulated as a \{\textit{verb, noun}\} pair. The combination of different \textit{verbs} and \textit{nouns} would lead to thousands of candidates~\cite{furnari2018leveraging}. Similar to the ``long-tailed'' distribution in many real-world applications, the majority of actions only occur very few times. Such imbalanced distribution would decrease the generalization capability of trained model on rare classes. In this report, we adopted a state-of-the-art method, Equalization Loss~\cite{tan2020equalization}, to handle the long-tailed distribution problem. 

\section{Methods}
We directly adopted the multi-modality feature provided by RULSTM~\cite{damen2020rescaling, furnari2020rolling}, which consists of features from three modalities, rgb $F_{rgb}$, flow $F_{flow}$, and object $F_{obj}$. $F_{rgb}$ and $F_{flow}$ were extracted from pretrained TSN models~\cite{wang2016temporal} on the action recognition task. $F_{obj}$ was formed by the object probability score predicted by pretrained FasterRCNN model~\cite{ren2015faster}. Each input $F\in\mathbb{R}^{N\times D_f}$ denotes the feature vector with a dimensionality of $D_f$ extracted from $N$ frames, (3.5-1)s before the beginning of the actions.

Our key idea is to exploit Transformer based attention mechanisms to fuse information from temporal dimension, different modalities, as well as verb/noun branches. The overall framework is illustrated in Fig.~\ref{fig:framework} and the details of each basic component are given below.  

\subsection{Temporal Self-Attention (TSA)}
Instead of applying conventional network architectures for temporally modelling like LSTM/GRU, we applied Transformer~\cite{vaswani2017attention} to better model the long-range temporal relationship by attention mechanisms. The input feature vector is added by sinusoidal positional embedding to incorporate the positional information. It transforms the input feature to a set of queries ($\mathbf{Q}$), keys ($\mathbf{K}$) and values ($\mathbf{V}$) via linear projection. Subsequently, the attention weights computed from the normalized dot product of $\mathbf{Q}$ and $\mathbf{K}$ are applied to aggregate values, as formulated in Eq.~\ref{eq:2}. It subsequently applies add \& norm operations to enable residual connections, as formulated in Eq.~\ref{eq:3}. Subsequently, non-linear feedforward MLPs followed by add \& norm residual connections are applied, as in Eq.~\ref{eq:4}.

\begin{equation}
\small
\mathbf{Q}=\mathbf{F}\mathbf{W}^q, \mathbf{K}=\mathbf{F}\mathbf{W}^k, \mathbf{V}=\mathbf{F}\mathbf{W}^v
\label{eq:1}
\end{equation}

where $\mathbf{W}^q\in\mathbb{R}^{D_f\times D_q}$, $\mathbf{W}^k\in\mathbb{R}^{D_f\times D_k}$, $\mathbf{W}^v\in\mathbb{R}^{D_f\times D_v}$ denote corresponding linear projection matrices.  

\begin{equation}
\small
\mathbf{A} = softmax\left (\frac{\mathbf{Q}\mathbf{K}^T}{\sqrt{D_k}}\right )\mathbf{V}
\label{eq:2}
\end{equation}

\begin{equation}
\small
\mathbf{F}^{'} = layer\_norm(\mathbf{A} + \mathbf{F}^{in})  
\label{eq:3}
\end{equation}

\begin{equation}
\small
\mathbf{F}^{out} = layer\_norm(\textbf{MLP}(\mathbf{F}') + \mathbf{F}^{'})  
\label{eq:4}
\end{equation}

\subsection{Cross-Modality Attention (CMA)}
To make use of the complementary information encoded in different modalities, we introduced a cross-modality attention (CMA) mechanism, which is expected to capture asynchronous yet relevant information across modalities. Inspired by the fusion method proposed in \cite{prakash2021multi}, we concatenate $F_{rgb}$ $F_{flow}$ $F_{obj}$ into a feature with a shape of $N\times \sum D_f$, and then apply the CMA module to aggregate features across time. 

\subsection{Symbiotic Attention (SA)}
Similar to previous action recognition/anticipation works, we utilized two branches to predict \textit{verb} and \textit{noun} separately. However, it is not appropriate to consider \textit{verb} and \textit{noun} as two independent variables to be predicted by two independent branches, since they share mutual contextual information \cite{wang2020symbiotic}. The awareness of the next active object provides the prior probability for predicting the next verb, whereas predicting the next verb would help recognize the next object to be manipulated. Therefore, we incorporated another Transformer module for the interaction between \textit{verb} and \textit{noun} branches. This module, referred to as Symbiotic Attention (SA) module, applied Transformer network to process concatenated feature input with a shape of $2N\times \sum D_f$. 

\subsection{Cascaded Architecture}
Based on the TSA, CMA, and SA modules, the illustration of our network architecture is given in Fig.~\ref{fig:framework}. It firstly processes the input of each modality by their corresponding TSA modules. Subsequently, the CMA modules in both branches fuse features across multiple modalities, followed by a SA module performing interactions between both branches. Finally, the features extracted from two branches are concatenated together and fed into another TSA module to predict the action.
We developed a cascaded architecture with the repetition of the same block, whereas the output of each block is extracted for prediction. In practice, the block number n is set as 2. 

\subsection{Equalization Loss}
To deal with the long tailed distribution, we adopted the Equalization Loss proposed in \cite{tan2020equalization}. It proposed a simple yet effective loss aimed at protecting the learning of rare classes by randomly neglecting the updating of rare classes when the target is a majority class. The loss function is modified from cross-entropy loss, and its formulation is shown as below,  
\begin{equation} \small
L_{SEQL}=-\sum_{j=1}^{c}y_{j}\log(\tilde{p}_{j})  
\end{equation}

\begin{equation} \small
\tilde{p}_{j}=\frac{e^{z_{j}}}{\sum_{k=1}^{c}\tilde{w}_{k}e^{z_{k}}}  
\end{equation}

\begin{equation} \small
\tilde{w}_{k}=1-\beta T_{\lambda}(y_{k})(1-y_{k}) 
\end{equation}
where $\beta$ is random binary variable with a probability of $\gamma$ to be 1 and otherwise 0. $T_{\lambda}(y_{k})$ is a threshold function determining whether $y_{k}$ is a majority class by predefined occurrence frequency threshold. 

\begin{table*}[h]
\small
\centering
 \caption{Results of Ablation Studies on Validation Set.}
  \begin{tabular}{@{}lccccccccc@{}}
   \toprule
       \multicolumn{1}{c}{\multirow{2}[0]{*}{Method}}  & \multicolumn{3}{c}{Overall (\%)}  & \multicolumn{3}{c}{Unseen (\%)}  & \multicolumn{3}{c}{Tail (\%)}  \\ \cmidrule(l){2-10} 
    & Verb & Noun & Action  & Verb & Noun & Action & Verb & Noun & Action  \\
    \midrule
    RULTSM\cite{damen2020rescaling}  & 27.76 & 30.76 & 14.04 & 28.78 & \underline{27.22} & \textbf{14.15} & 19.77 & 22.02 & 11.14 \\
    \midrule  
    TSA-RGB  & 33.23 & \underline{32.65} & 13.71 & 28.65 & 20.61 & 10.23 & 29.12 & \underline{31.41} & 13.34 \\
    TSA-Flow & 24.19 & 17.02 & 6.74 & 30.61 & 15.74 & 6.01 & 19.33 & 15.46 & 5.72 \\
    TSA-Obj  & 25.37 & 29.51 & 9.93 & 28.39 & 22.19 & 7.06 & 21.26 & 28.09 & 9.51 \\
    w/o CMA & 31.46 & 31.92 & 14.90 & \underline{34.10} & 23.47 & 10.22 & 26.37 & 30.14 & \underline{14.56} \\
    w/o SA  & \textbf{35.78} & 32.18 & 12.93 & 29.79 & 17.56 & 10.51 & \textbf{32.08} & 31.01 & 12.43 \\
    w/o Equal & 27.65 & 31.34 & 14.16 & 27.49 & 25.25 & 12.61 & 20.92 & 25.60 & 11.98 \\
    \midrule
    Proposed-Single  & 33.60 & 32.54 & \underline{15.05} & 33.05 & 25.43 & 11.96 & 29.04 & 31.03 & 14.39 \\
    Proposed-Ensemble & \underline{35.04} & \textbf{35.49} & \textbf{16.60} & \textbf{34.64} & \textbf{27.26} & \underline{13.83} & \underline{30.08} & \textbf{33.64} & \textbf{15.53} \\
    \bottomrule
    \end{tabular}
    \label{tab:ablation}
\end{table*}

\begin{table*}[h]
 \small
  \centering
  \caption{Results of Testing Set on LeaderBoard.}
    \begin{tabular}{@{}lccccccccc@{}}
   \toprule
   \multicolumn{1}{c}{\multirow{2}[0]{*}{Method}}  & \multicolumn{3}{c}{Overall (\%)}  & \multicolumn{3}{c}{Unseen (\%)}  & \multicolumn{3}{c}{Tail (\%)}  \\ \cmidrule(l){2-10} 
    & Verb & Noun & Action & Verb & Noun & Action & Verb & Noun & Action  \\ 
    \midrule
     RULSTM-RGB    & 24.69  & 26.38  & 10.45  & 17.88  & 23.16  & 9.13  & 17.32  & 16.79  & 7.39 \\
     RULSTM-Flow   & 21.24  & 18.12  & 7.36   & 17.27  & 18.95  & 6.86  & 13.54  & 9.44   & 4.97 \\
     RULSTM-OBJ    & 13.93  & 15.17  & 3.96   & 14.05  & 20.41  & 5.79  & 6.18   & 5.37   & 1.85 \\
     RULSTM-Fusion & 25.25  & 26.69  & 11.19  & 19.36  & 26.87  & 9.65  & 17.56  & 15.97  & 7.92 \\
     Proposed-Single    & 37.13  & 30.19  & 12.44  & 29.72  & 20.87  & 10.57  & 34.53  & 28.42  & 9.74 \\
     Proposed-Ensemble  & 36.15  & 32.20  & 13.39  & 27.60  & 24.24  & 10.05 & 32.06  & 29.87  & 11.88 \\
    \bottomrule
    \end{tabular}
  \label{tab:result}
\end{table*}

\section{Implementation Details}
The whole model was implemented with Pytorch and trained on a single RTX 2080 Ti GPU. The batch size was set as 128 and we applied SGD optimizer with a learning rate of 0.01 and a momentum of 0.9.     
The implementation details can be found in \url{https://github.com/guxiao0822/trans_action}. 

To participate in the challenge, we developed an ensemble of three trained models based on our proposed method together with the baseline {RULSTM-Fusion} to achieve performance gains from their complementary information. 
\section{Results and Discussion}
Following the evaluation guideline of this challenge\footnote{\url{https://competitions.codalab.org/competitions/25925}}, the Mean Top-5 Recall Metric is used. First of all, to demonstrate the effectiveness of different modules proposed, we conducted ablation study on the validation subset with the results shown in Table~\ref{tab:ablation}. The TSA-RGB/Flow/Obj refers to the variant only applying TSA with their corresponding single-modality feature as input. w/o CMA, SA denote the variants with CMA, SA module removed respectively. w/o Equal replaces the Equalization Loss by the conventional cross-entropy loss. It can be observed that overall the complete method performs well. 

For the test set, The final results of our single model and the ensemble version are given in Table~\ref{tab:result}, together with the results of the baseline method RULSTM~\cite{furnari2020rolling}. As shown in Table~\ref{tab:result}, for our single model, our method competes against the baseline methods regarding most metrics. Especially for the tail classes, a significant improvement can be observed. The ensemble of our models and RULSTM\_Fusion leads to slight improvement in terms of some metrics, especially for the result of Tail \textit{action}. It is also noteworthy that our proposed method ranked 1st for \textit{verb} in all three (sub)sets.      
  
We noticed marginally preferable results reported by some other teams in terms of \textit{action} as shown in the Leaderboard. Future work should be targeted at further exploring the symbiotic relationship between \textit{verb} and \textit{noun} for the improvement of \textit{action} classification. Modelling the temporal transition of different actions as well as the label distribution to handle label uncertainty should also be taken into consideration.

{\small
\bibliographystyle{ieee}
\bibliography{paper}
}

\end{document}